\documentclass[runningheads]{llncs}

\usepackage{eccv}

\usepackage{eccvabbrv}

\usepackage{graphicx}
\usepackage{booktabs}

\usepackage[accsupp]{axessibility}  % Improves PDF readability for those with disabilities.

\usepackage[pagebackref,breaklinks,colorlinks]{hyperref}
\usepackage{orcidlink}
\usepackage{graphicx}
\usepackage{caption}
\usepackage{amsmath,amssymb} % define this before the line numbering.
\usepackage{bm}
\usepackage{multirow}
\usepackage{pifont}
\usepackage{hyperref}
\usepackage{color}
\usepackage{amssymb}
\usepackage{booktabs}

\begin{document}

\title{Like Humans to Few-Shot Learning through Knowledge Permeation of Vision and Text}

\titlerunning{Few-Shot Learning through Knowledge Permeation of Vision and Text}

\newcommand{\equalcontrib}{\textsuperscript{*}}

\author{Yuyu Jia \inst{1} \equalcontrib \orcidlink{0009-0002-2303-8995} \and
Qing Zhou\inst{1} \equalcontrib \and Wei Huang\inst{2} \and Junyu Gao\inst{1} \and Qi Wang\inst{1}}

\authorrunning{Y.Jia et al.}

\institute{School of Artificial Intelligence, Optics and Electronics (iOPEN), Northwestern Polytechnical University, Xi’an 710072, Shaanxi, P. R. China\\
\email{\{jyy2019,mrazhou\}@mail.nwpu.edu.cn, gjy3035@gmail.com, crabwq@gmail.com}\\ 
\and
Data Science in Earth Observation, Technical University of Munich, Munich 80333, Germany\\
\email{hw2hwei@gmail.com}}

\maketitle

\renewcommand{\thefootnote}{\fnsymbol{footnote}}
\footnotetext[1]{Equal contribution.}
\renewcommand{\thefootnote}{\arabic{footnote}}

\begin{abstract}
Few-shot learning aims to generalize the recognizer from seen categories to an entirely novel scenario.
With only a few support samples, several advanced methods initially introduce class names as prior knowledge for identifying novel classes.
However, obstacles still impede achieving a comprehensive understanding of how to harness the mutual advantages of visual and textual knowledge.
In this paper, we propose a coherent \textbf{Bi}directional \textbf{K}n\textbf{o}wledge \textbf{P}ermeation strategy called \textbf{BiKop}, which is grounded in a human intuition: A class name description offers a \textit{general} representation, whereas an image captures the \textit{specificity} of individuals.
BiKop primarily establishes a hierarchical joint general-specific representation through bidirectional knowledge permeation.
On the other hand, considering the bias of joint representation towards the base set, we disentangle base-class-relevant semantics during training, thereby alleviating the suppression of potential novel-class-relevant information.
Experiments on four challenging benchmarks demonstrate the remarkable superiority of BiKop.
Our code will be publicly available.
  
\end{abstract}

\section{Introduction}
As a forefront area of research in computer vision, Few-Shot Learning (FSL) investigates how models can rapidly generalize existing knowledge to novel scenarios, mirroring human-like adaptability.
It first \textbf{acquires a base classifier through pre-training} with ample samples, then \textbf{learns to identify unseen/novel classes with a few support samples} \cite{snell2017prototypical}, \cite{Finn_Abbeel_Levine_2017}, \cite{Ravi_Larochelle_2017}, \cite{jia2023exploring}.
In this setting, two significant challenges naturally arise:
(1) the collapse of sparse feature representation induced by significant intra-class variations,
(2) the bias toward the base set during training might compromise the model's fragile generalization.
To tackle these issues, previous FSL methods primarily concentrate on the adaptive optimization of meta-learners \cite{Finn_Abbeel_Levine_2017}, \cite{Nichol_Achiam_Schulman_2018}, \cite{Rusu_Rao_Sygnowski_Vinyals_Pascanu_Osindero_Hadsell_2018}, 
the effective mining of prior knowledge, \cite{hao2023class}, \cite{Doersch_Gupta_Zisserman_2020}, \cite{Mangla_Singh_Sinha_Kumari_Balasubramanian_Krishnamurthy_2020}, 
or the formulation of judicious metric functions \cite{Vinyals_Blundell_Lillicrap_Kavukcuoglu_Wierstra_2016}, \cite{snell2017prototypical}, \cite{Zhang_Cai_Lin_Shen_2020}.

\begin{figure}[h]
    \centering
    \includegraphics[width=0.95\textwidth]{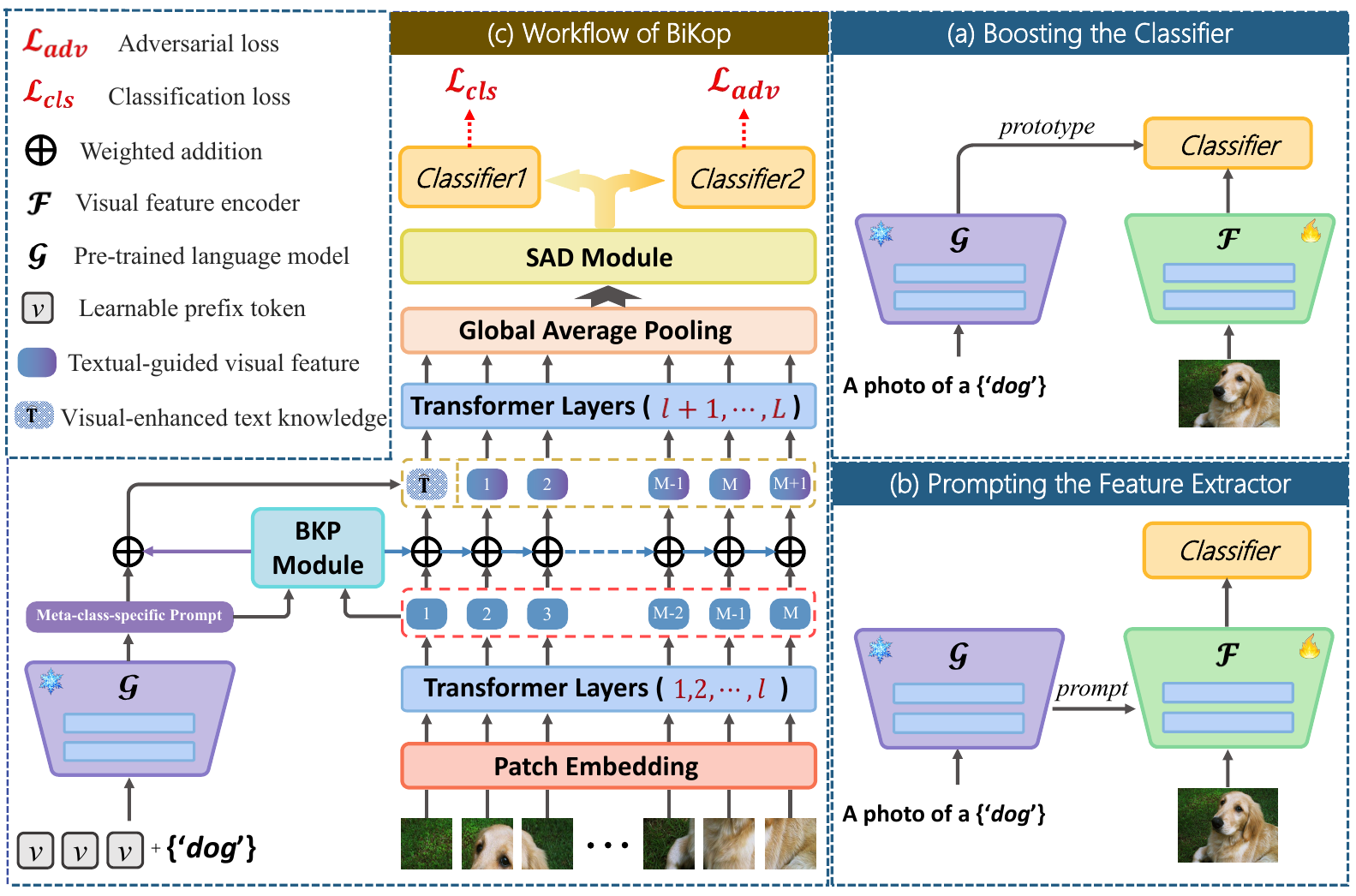}
    \caption{Comparison of BiKop with studies of introducing textual knowledge.
    \textcolor{blue}{\textbf{(a)}} They embed class names into text prototypes, directly employed to enhance classifiers.
    \textcolor{blue}{\textbf{(b)}} Several recent methods utilize textual information to modulate the extraction of visual features.
    \textcolor{blue}{\textbf{(c)}} Workflow of our BiKop. To alleviate the collapse of sparse feature representation, the BKP module harnesses the complementary advantages between textual and visual knowledge by the bidirectional permeation of both. Furthermore, the SAD module adversarially disentangles the base-class-relevant semantic to mitigate the base set bias and boost the model’s generalization to novel categories.} 
    \label{fig_1}
\end{figure}

Driven by advancements in Natural Language Processing (NLP) models \cite{Devlin_Chang_Lee_Toutanova_2019}, \cite{Radford_Narasimhan_Salimans_Sutskever}, \cite{pennington2014glove}, \cite{Radford_Kim_Hallacy_Ramesh_Goh_Agarwal_Sastry_Askell_Mishkin_Clark_et_al}, class name embeddings as a direct form of prior knowledge are injected to enhance the few-shot recognizer \cite{Xing_Rostamzadeh_Oreshkin_Pinheiro_2019}, \cite{Yan_Bouraoui_Wang_Jameel_Schockaert_2021} as illustrated in Fig. \ref{fig_1}(a).
Not confined to this, some recent studies propose that leveraging textual information to modulate the extraction of visual features can unlock further potential (Fig. \ref{fig_1}(b).
For example, SP \cite{chen2023semantic} utilizes textual information as prompts to fine-tune visual features.
CMA \cite{lin2023multimodality} repurposes class names as additional one-shot training samples, implicitly intervening in visual features.

While these methods have exhibited considerable promise, there are still crucial challenges.
\textbf{(L1)} The effective utilization of mutual knowledge between vision and text has not been achieved.
Typically, a class name offers a general representation; an image encapsulates the specificity of individuals.
For instance, given a sentence like ``\textit{a photo of the dog}'', we evoke the \textbf{general} concept of ``dog'' and its shared characteristics, while an image concretely characterizes the individual \textbf{specificity} of the category of ``dog''.
In this spirit, we posit that explicitly constructing the general-specific hierarchical knowledge structure can effectively alleviate the collapse of sparse feature representation.
\textbf{(L2)} Although textual knowledge brings gains, the inclusion of class names as base-class-relevant information during training could exacerbate the model bias towards the base set, which is an inherent challenge in FSL\cite{Fan_Tang_Tai}, \cite{Ma_Sun_Yang_Yang}, \cite{Luo_Xu_Xu_2022}.
In other words, such gains may be significantly compromised when dealing with novel classes.
This previously overlooked contradiction undermines the enhancements facilitated by textual knowledge.

In this paper, we propose BiKop, a novel approach comprising two modules, Bidirectional Knowledge Permeation (BKP) and Semantic Adversarial Disentanglement (SAD), to jointly mitigate the representation collapse (L1) and model bias (L2). 
BKP allows the bidirectional permeation of mutual knowledge derived from textual and visual modalities.
It involves three steps.
Firstly, inspired by the use of class names as prompts in \cite{chen2023semantic}, we creatively extend it to meta-class-specific prompts to encapsulate textual knowledge with stronger meta-task adaptability.
Secondly, a lightweight cross-attention block is introduced to enable the permeation of general knowledge, from textual prompts to visual features.
In the opposite direction, individual specificity knowledge from vision features is permeated into textual prompts to enhance their diversity.
Thirdly, the enhanced prompts and visual features are fed into a Transformer encoder, producing a robust joint feature representation for sparse support samples.
On the other hand, we consider that introducing base-class-relevant knowledge (\textit{i.e.,} class names) exacerbates the bias towards the base set.
In other words, the model inevitably suppresses the potential semantics of novel categories during training to achieve better convergence.
In response to this, SAD adversarially disentangles base-class-relevant semantics from the joint feature representation, which balances the model's convergence on the base set and generalization to novel classes.

By combining these two modules, BiKop not only gains advantages from the BKP module to improve robustness in sparse feature representation but also alleviates the interference of base-class-relevant knowledge on the model's generalization.
Through exhaustive experiments conducted across four benchmarks, BiKop consistently manifests performance enhancements with varying backbones.
Particularly, with the popular backbones of Visformer-T\cite{Chen_Xie_Niu_Liu_Wei_Tian_2021} and ViT-S\cite{Dosovitskiy_Beyer_Kolesnikov_Weissenborn_Zhai_Unterthiner_Dehghani_Minderer_Heigold_Gelly_et_al}, the 1-shot classification accuracy on the miniImageNet dataset has been improved by 5.76\% and 7.58\%, respectively.

In summary, the contribution of this paper is threefold:

\begin{enumerate}
     \item[--] We take a close look at the complementary disparities within textual and visual knowledge, ingeniously combining the two to construct a novel few-shot image recognition method named BiKop.

     \item[--] The proposed BKP module extends class names as meta-class-specific prompts and conducts bidirectional permeation of textual and visual knowledge, fully harnessing the mutual advantages of both to alleviate the collapse of sparse feature representation.
     Besides, the SAD module mitigates the base set bias exacerbated by base-class-relevant knowledge (\textit{i.e.,} class names), further enhancing the model's generalization to novel categories.
     
     \item[--] Our proposed method is easy to implement, and seamlessly compatible with different backbone architectures.
     Through extensive experiments, we validate that BiKop achieves state-of-the-art performance across different FSL datasets.
\end{enumerate}

\section{Related Work}
\label{sec:related work}

\textbf{Few-shot Classification.}
Over the past few years, a persistent fervor surrounds Few-Shot Learning (FSL) as scholars ardently seek to enhance the generalization of artificial intelligence models.
Among the significant applications of FSL, few-shot classification can be broadly categorized into two groups: metric-based methods \cite{Vinyals_Blundell_Lillicrap_Kavukcuoglu_Wierstra_2016}, \cite{snell2017prototypical}, \cite{Fort_2018}, \cite{Ma_Fang_Avraham_Zuo_Drummond_Harandi}
and optimisation-based methods \cite{Finn_Abbeel_Levine_2017}, \cite{Nichol_Achiam_Schulman_2018}, \cite{Oh_Yoo_Kim_Yun_2020}, \cite{Zintgraf_Shiarlis_Kurin_Hofmann_Whiteson_2018}, \cite{Jelley_Storkey_Antoniou_Devlin_2023}.
Metric-based approaches concentrate on acquiring a feature space, in which a suitable distance function is employed for measuring similarity. 
Optimization-based methods conduct rapid adaption with a few training samples for novel categories through learning a good meta-optimizer.
Additionally, recent self-supervised-based methods \cite{hiller2022rethinking}, \cite{hao2023class}, \cite{Mangla_Singh_Sinha_Kumari_Balasubramanian_Krishnamurthy_2020}, \cite{Lu_Wen_Liu_Liu_Tian_2022}, \cite{Yang_Wang_Zhu_2022} have demonstrated their potential for exploring in few-shot settings.
They fundamentally leverage pretext tasks on the base set to extract more prior knowledge, thereby benefiting the model's generalization.

\noindent
\textbf{Few-shot Learning with Textual Knowledge.}
Instead of mining prior or generalizing knowledge from training on the base set, a series of advanced investigations \cite{Akyrek_Akyrek_Wijaya_Andreas_2021}, \cite{Li_Huang_Lan_Feng_Li_Wang_2020}, \cite{Peng_Li_Zhang_Li_Qi_Tang_2019}, \cite{Xing_Rostamzadeh_Oreshkin_Pinheiro_2019} directly incorporate prior information from other modalities, especially textual knowledge, to facilitate the recognition of novel classes.
For example, an adaptive fusion mechanism is introduced in \cite{Xing_Rostamzadeh_Oreshkin_Pinheiro_2019} to merge a visual prototype with a semantic prototype derived from the class name embedding.
In \cite{Peng_Li_Zhang_Li_Qi_Tang_2019}, the additional textual knowledge and visual information are effectively integrated to infer desired few-shot classifiers.
SP \cite{chen2023semantic} treats textual knowledge as prompts to modulate the extraction of visual features, extending its focus beyond just optimizing classifiers.
Different from prior studies that utilize textual knowledge at the level of classifiers or with rigid prompts, we implement bidirectional knowledge permeation to harness the mutual advantages of textual and visual modalities for sparse feature representation.

\noindent
\textbf{Prompt Learning.}
With the support of massive data, large-scale visual-language models \cite{Anderson_Wu_Teney_Bruce_Johnson_Sunderhauf_Reid_Gould_van}, \cite{Antol_Agrawal_Lu_Mitchell_Batra_Zitnick_Parikh_2015}, \cite{Xu_Ba_Kiros_Cho_Courville_Salakhudinov_Zemel_Bengio_2015}, \cite{Qiao_Zhang_Xu_Tao_2019} have witnessed rapid development in recent years.
However, in the face of diverse downstream tasks, efficiently adapting large models with limited data has become a new research focus.
Prompt learning, introduced from natural language processing into visual tasks, adapts to various downstream tasks by optimizing prompts with few parameters instead of tuning deep models.
For example, the manually crafted template ``a photo of a [CLASS]" in CLIP \cite{Radford_Kim_Hallacy_Ramesh_Goh_Agarwal_Sastry_Askell_Mishkin_Clark_et_al} is employed to represent the textual embedding for zero-shot prediction.
Considering the difficulty of learning task-specific knowledge with manually crafted template prompts, Context Optimization (CoOp) \cite{Zhou_Yang_Loy_Liu_2022} introduces learnable tokens to construct soft prompts.
Conditional Context Optimization (CoCoOp) \cite{Zhou_Yang_Loy_Liu_D} is proposed to learn instance-specific prompts for further generalization to unseen categories.
Comparatively, our BiKop makes the first attempt to manage meta-class-specific prompts for adapting prompts to the meta-learning scheme, achieving meta-task-specific optimization of textual knowledge for each category.
% 适应元学习框架，为每个文本知识增加元任务相关性，来提升分类性能（之后可以加实验）

\section{Problem Definition}
\label{sec:problem_definition}
Few-shot classification generalizes the knowledge learned on base classes $\mathcal{C}_{base}$ with abundant labeled data to sparse samples from novel classes $\mathcal{C}_{novel}$, where $\mathcal{C}_{base} \cap \mathcal{C}_{base}= \emptyset$.
We adopt the popular episodic manner \cite{Vinyals_Blundell_Lillicrap_Kavukcuoglu_Wierstra_2016}, \cite{snell2017prototypical} to define the $N$-way $K$-shot classification task, where $N$ represents the number of classes, and $K$ represents the number of labeled images per class.
In each episode, we are given support set $\mathcal{S} =\left \{ (x_{i}^{s},y_{i}^{s}) \right\}_{i=1}^{NK}$ and query set $\mathcal{Q} =\left \{ (x_{i}^{q},y_{i}^{q}) \right\}_{i=1}^{NQ}$, where $Q$ denotes the number of images to be classified for each category.
We aim to acquire a model with good generalization to novel classes after meta-training on the base set.

\section{Bidirectional Knowledge Permeation and Semantic Disentanglement (BiKop)}

\subsection{Design guideline of network structure}
\label{sec:overview}

\noindent
\textbf{Preparation.} 
Learning a generalized feature extractor has been proven to be beneficial for few-shot classification.
Thus, our approach involves two stages: pre-training and fine-tuning.
For the sake of ensuring a fair comparison, we followed consistent pre-training procedures with \cite{hao2023class} and \cite{chen2023semantic} for different feature extractors (\textit{i.e.,} Visformer-T \cite{Chen_Xie_Niu_Liu_Wei_Tian_2021} and ViT-S \cite{Dosovitskiy_Beyer_Kolesnikov_Weissenborn_Zhai_Unterthiner_Dehghani_Minderer_Heigold_Gelly_et_al}). Specific details can be referred to the aforementioned literature.
We focus on utilizing BiKop for fine-tuning the feature extractor, addressing the challenges of sparse feature representation collapse and model bias, thereby improving the model's performance in scenarios with scarce data.
Fig. \ref{fig_1}(c) illustrates the pipeline of BiKop.
The input image $\bm{x} \in \mathbb{R}^{H \times W\times C}$ ($H$, $W$, $C$ are the height, width, and dimension) is divided into $M=H\cdot W/P^{2}$ image patches, with each patch $p^{i} \in \mathbb{R}^{P^{2}\times C}$.
Then, all flattened patches of support and query images are fed into $L$-layers Transformer architecture $\mathcal{F}(\cdot)$ to extract visual features.

\vspace{\baselineskip}
\noindent
\textbf{The collapse of sparse feature representation.} 
The quality of the supporting feature representation determines the performance of the few-shot classifier.
However, due to significant intra-class variations, a minimal number of support samples makes it challenging to establish a generalizable and discriminative feature representation.
To mitigate this issue, we establish a hierarchical joint general-specific representation through bidirectional knowledge permeation, maximizing the mutual advantages of visual and textual modalities.
Concretely, given a support image $\bm{x}^{s}$, 
the \textbf{B}idirectional \textbf{K}nowledge \textbf{P}ermeation (BKP) module (sec.\ref{sec:BKP}) takes its visual features at the $l^{th}$ layer $\bm{Z}_{l}^{s}$ and its class name $y^{text}$ as inputs, producing textual-guided visual features $\bm{\hat{Z}}_{l}^{s}$ and visual-enhanced textual knowledge $\mathbf{\hat{T}}$:
\begin{equation}
\label{eqt_1}
\bm{\hat{Z}}_{l}^{s}, \mathbf{\hat{T}} = \mathcal{BKP}(\bm{Z}_{l}^{s}, y^{text}),
\end{equation}
where $\bm{\hat{Z}}_{l}^{s}, \bm{Z}_{l}^{s} \in \mathbb{R}^{M \times C_{d}}$, $\mathbf{\hat{T}} \in \mathbb{R}^{C_{d}}$, and $C_{d}$ is the number of output feature channels.
We utilize $\mathbf{\hat{T}}$ as prompts to modulate $\bm{\hat{Z}}_{l}^{s}$ through the remaining Transformer layers, in which the Multi-head Self-Attention (MSA) mechanisms allow the interaction between prompts and visual features. 
Then, we construct the robust joint feature representation for sparse support samples as:
\begin{equation}
\label{eqt_2}
\mathbf{F}^{s} = GAP(MSA([\bm{\hat{Z}}_{l}^{s}; \mathbf{\hat{T}}])),
\end{equation}
where $\mathbf{F}^{s} \in \mathbb{R}^{C_{d}}$, $[\cdot;\cdot]$ denotes the concatenation operation, and $GAP(\cdot)$ stands for the Global Average Pooling function.
Subsequently, the robust feature representation is averaged within each category to compute the class prototypes:
\begin{equation}
\label{eqt_3}
\bm{p}_{c}=\frac{1}{K}\sum_{k=1}^{K}\mathbf{F}^{s|c}_{k},
\end{equation}
where $\bm{p}_{c} \in \mathbb{R}^{C_{d}}$ is the prototype of the category $c$, and $\mathbf{F}^{s|c}_{k}$ denotes the $k^{th}$ support feature of the category $c$.

\vspace{\baselineskip}
\noindent
\textbf{The model bias exacerbated by base-class-relevant knowledge.}
For a specific class $c$ in the base set, despite its prototype $\bm{p}_{c}$ incorporating hierarchical knowledge of both general and specific aspects, introducing base-class-relevant knowledge from class names inevitably exacerbates the bias toward the base set.
Specifically, we assume that $\bm{p}_{c} = \textit{Mix}(\bm{\hat{p}}_{c}, \bm{\check{p}}_{c})$ is constructed from a mix of class-relevant semantics $\bm{\hat{p}}_{c}$ and class-irrelevant semantics $\bm{\check{p}}_{c}$.
During training, the model suppresses the learning of $\bm{\check{p}}_{c}$ to achieve better convergence to base classes.
This undermines the model's representational capacity for the potential semantics of novel categories.

For a better trade-off between model convergence and model generalization, the \textbf{S}emantic \textbf{A}dversarial \textbf{D}isentanglement (SAD) module (sec.\ref{sec:SAD}) adversarially disentangles base-class-relevant semantics $\bm{\hat{p}}_{c}$ and releases the suppression of base-class-irrelevant semantics $\bm{\check{p}}_{c}$:
\begin{equation}
\label{eqt_4}
\bm{\hat{p}}_{c}, \bm{\check{p}}_{c}=\mathcal{SAD}(\bm{p}_{c}).
\end{equation}
% Note that the SAD module is not involved in the inference process.
During training, the pre-trained language model $\mathcal{G}(\cdot)$ is fixed and other parameters are optimized by \textit{maximizing} the similarities between query features and $\bm{\hat{p}}_{c}$ with a cross-entropy loss:
\begin{equation}
\label{eqt_5}
\mathcal{L}_{cls} = -\mathbb{E}_{S,Q}\mathbb{E}_{\bm{x}^{q}}log\frac{\mathrm{exp}(\delta(\mathbf{F}(\bm{x}^{q}),\bm{\hat{p}}_{y^{q}})/\tau)}
{{\textstyle \sum_{i=1}^{N}}\mathrm{exp}(\delta(\mathbf{F}(\bm{x}^{q}),\bm{\hat{p}}_{i})/\tau)},
\end{equation}
where $\bm{\hat{p}}_{y^{q}}$ is the base-class-relevant semantics of category $y^{q}$, $\tau$ is a temperature hyper-parameter, and $\delta(\cdot)$ represents the cosine similarity.
Furthermore, we \textit{minimize} the similarities between query features and their base-class-irrelevant semantics $\bm{\check{p}}_{y^{q}}$, thereby liberating the model for respresenting potential novel class semantics.
\begin{equation}
\label{eqt_6}
\mathcal{L}_{adv} = \mathbb{E}_{S,Q}\mathbb{E}_{\bm{x}^{q}}log\frac{\mathrm{exp}(\delta(\mathbf{F}(\bm{x}^{q}),\bm{\check{p}}_{y^{q}})/\tau)}
{{\textstyle \sum_{i=1}^{N}}\mathrm{exp}(\delta(\mathbf{F}(\bm{x}^{q}),\bm{\check{p}}_{i})/\tau)}.
\end{equation}

To be clear, the overall loss of BiKop is obtained by weighted summation:
\begin{equation}
\label{eqt_7}
\mathcal{L}_{total} = \mathcal{L}_{cls} + \gamma \cdot \mathcal{L}_{adv}.
\end{equation}

\begin{figure}[t]
  \begin{minipage}[t]{0.48\textwidth}
    \centering
    \includegraphics[height=5.7cm]{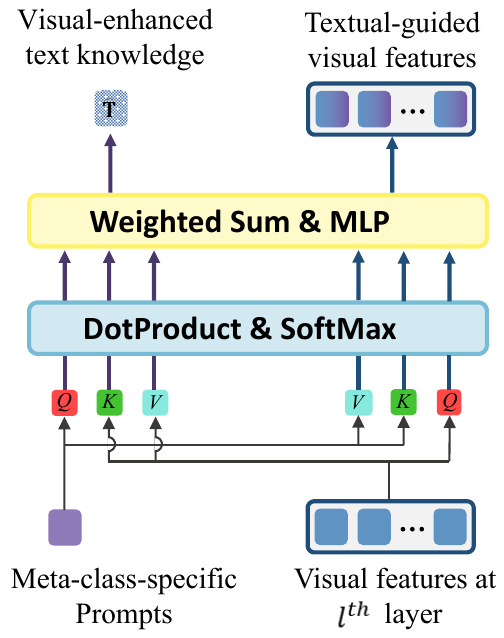}
    \captionsetup{width=0.9\linewidth, justification=justified}
    \caption{Configuration diagram of the Bidirectional Knowledge Permeation (BKP) module.}
    \label{fig_2}
  \end{minipage}%
  \begin{minipage}[t]{0.52\textwidth}
    \centering
    \includegraphics[height=5.7cm]{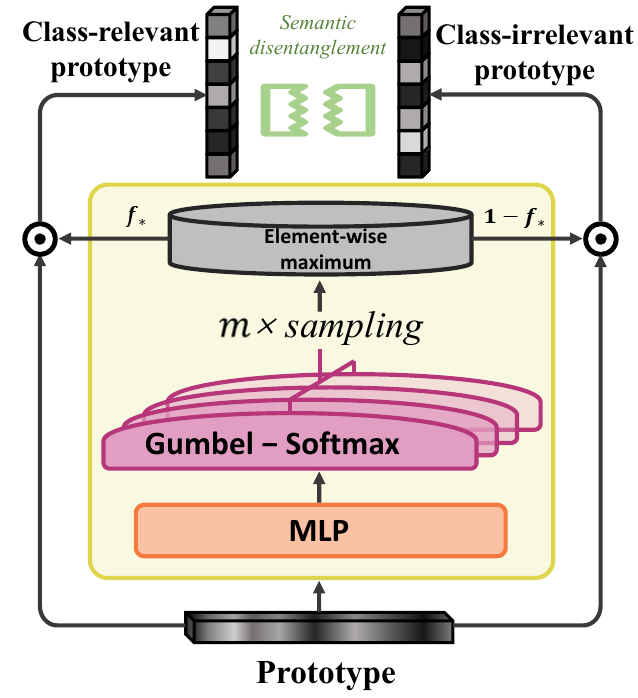}
    \captionsetup{width=0.9\linewidth, justification=justified}
    \caption{Visualization for the implementation of the Semantic Adversarial Disentanglement (SAD) module.}
    \label{fig_3}
  \end{minipage}
\end{figure}

\subsection{BKP Module}
\label{sec:BKP}
As shown in Fig. \ref{fig_2}, given the meta-class-specific prompts and the $l^{th}$-layer visual features, the BKP module $\mathcal{BKP}(\cdot)$ conducts bidirectional permeation of the textual and visual knowledge.

\noindent
\textbf{Meta-class-specific Prompts.}
Different from the prior FSL study \cite{chen2023semantic} that directly employs class name embeddings as prompts, we further extend it to meta-class-specific prompts for better adaptation to the meta-learning scheme. 
Experiments in sec. \ref{meta-class-specific prompt} prove the efficacy of this design.
In an episode containing $N$ categories, we use $w$ learnable prefix tokens to create a meta-class-specific template for each category. 
Accordingly, the $\left \{class \ name \right\}$, \textit{i.e.,} $y^{text}$ is filled into the corresponding template:
\begin{equation}
\label{eqt_8}
\mathbf{E}_{n} = [v]_{1}^{n}[v]_{2}^{n}\dots [v]_{w}^{n}[class \ name]^{n}, \ \ \ \ \ n=1, \dots N,
\end{equation}
where $[v]_{i}^{n}, i\in \left \{ 1,2,\dots,w \right\}$ denotes the learnable prefix tokens of $n^{th}$ category, $[class \ name]^{n}$ represents the $n^{th}$ class name.
Each meta-class-specific prompt $\mathbf{E}_{n}$ is then fed into a pre-trained CLIP text encoder $\mathcal{G}(\cdot)$ to extract the corresponding prompt embedding $\mathbf{T}_{n}= \mathcal{H}(\mathcal{G}(\mathbf{E}_{n}))$ ($\mathcal{H}$ projects the prompt to the same dimension $C_{d}$ of visual features).

\noindent
\textbf{Permeating Textual Knowledge into Visual Knowledge.}
It aims to supplement the lack of general knowledge in sparse visual features.
Given the support visual features from the $l^{th}$ Transformer layer $\bm{Z}_{l}^{s} \in \mathbb{R}^{M \times C_{d}}$ and prompt embedding $\mathbf{T}\in \mathbb{R}^{1 \times C_{d}}$ for its corresponding category, we get a visual-to-text similarity map $\mathbf{A}_{\bm{Z}}\in \mathbb{R}^{M \times 1}$:
\begin{equation}
\label{eqt_9}
\mathbf{A}_{\bm{Z}} = SoftMax((\bm{Z}_{l}^{s}W_{q})\otimes (\mathbf{T}W_{k})^{\top}/\sqrt{C_{d}}).
\end{equation}
The textual-guided visual feature is then formulated as:
\begin{equation}
\label{eqt_10}
\bm{\hat{Z}}_{l}^{s} = \bm{Z}_{l}^{s} + \mu \cdot \mathcal{X}(\mathbf{A}_{\bm{Z}}\otimes \mathbf{T}W_{v}),
\end{equation}
where $\mu$ is the weight coefficient, $\otimes$ denotes matrix multiplication, and $\mathcal{X}$ is a two-layer MLP \cite{Tolstikhin_Houlsby_Kolesnikov_Beyer_Zhai_Unterthiner_Yung_Steiner_Keysers_Uszkoreit_et} block. $W_{q}\in \mathbb{R}^{C_{d} \times C_{d}}$, $W_{k}\in \mathbb{R}^{C_{d} \times C_{d}}$, and $W_{v}\in \mathbb{R}^{C_{d} \times C_{d}}$ are 
linear transformation parameters.

\noindent
\textbf{Permeating Visual Knowledge into Textual Knowledge.}
In this process, visual features enrich the individual specificity of textual knowledge, \textit{i.e.,} prompt embedding.
Similarly to Eqs. \ref{eqt_9} and \ref{eqt_10}, using a shared structure and parameters, the visual-enhanced textual knowledge can be written as:
\begin{align}
\label{eqt_11_12}
\mathbf{A}_{\mathbf{T}} &= SoftMax((\mathbf{T}W_{q})\otimes (\mathbf{Z}_{l}^{s}W_{k})^{\top}/\sqrt{C_{d}}) \\ \mathbf{\hat{T}} &= \mathbf{T} + \mu \cdot \mathcal{X}(\mathbf{A}_{\mathbf{T}}\otimes \mathbf{Z}_{l}^{s}W_{v}).
\end{align}

\subsection{SAD Module}

\label{sec:SAD}
Inspired by \cite{Luo_Xu_Xu_2022}, which advocates that channel importance matters model bias, we propose the SAD module $\mathcal{SAD}(\cdot)$ as shown in Fig. \ref{fig_3}, to adversely disentangle base-class-relevant and base-class-irrelevant semantics from the perspective of channel representations, \textit{i.e.,} prototypes.
Then, they are used to release the suppression of potential novel class semantics through an adversarial optimization strategy, as in Eq. \ref{eqt_6}.

We design a differentially samplable filter $\bm{f}_{*} \in \mathbb{R}^{C_{d}}$ for each category to learn the contribution of each dimension of the prototype.
In this way, the class-relevant and class-irrelevant prototypes can be written as:
\begin{equation}
\label{eqt_13}
\bm{\hat{p}}_{*} = \bm{p}_{*} \odot \bm{f}_{*}, \bm{\check{p}}_{*} = \bm{p}_{*} \odot (1 - \bm{f}_{*}),
\end{equation}
where $\odot$ denotes element-wise multiplication.
To approximate the optimal differentially samplable filter $\bm{f}_{*}$, we apply the Gumbel-Softmax trick \cite{Jang_Gu_Poole_2016}, \cite{Lv_Liang_Li_Zang_Liu_Wang_Liu}, \cite{Chen_Song_Wainwright_Jordan_2018} to weighted sample a $C_{d}$-dimensional random vector independently for $m$ times:
\begin{equation}
\label{eqt_14}
O_{j} = \mathrm{Gumbel-Softmax}(\mathcal{D}(\bm{p}_{*})), \ \ \ j=1,2,\dots m,
\end{equation}
where $\mathcal{D}(\cdot)$ is an MLP block that maps the input to a $C_{d}$-dimensional vector.
Then, $\bm{f}_{*}$ is accquired from the element-wise maximum of $O_{1},O_{2},\dots,O_{m}$:
\begin{equation}
\label{eqt_15}
\bm{f}_{*} = (f^{1}, f^{2}, \dots, f^{C_{d}}), \ \ \ f^{i}=\underset{j}{\mathrm{max}}O_{j}^{i},
\end{equation}
where $f^{i}, i\in \left \{ 1,2,\dots,C_{d} \right\} $represents the $i^{th}$ dimension of $\bm{f}_{*}$, $O_{j}^{i}$ stands for the $i^{th}$ dimension of the vector obtained in the $j^{th}$ sampling.

\section{Experiments}

\subsection{Datasets}
\label{sec:dataset}
Our method is evaluated on four widely used few-shot classification benchmarks: \textit{mini}ImageNet \cite{Vinyals_Blundell_Lillicrap_Kavukcuoglu_Wierstra_2016}, \textit{tiered}ImageNet \cite{Ren_Triantafillou_Ravi_Snell_Swersky_Tenenbaum_Larochelle_Zemel}, CIFAR-FS \cite{Bertinetto_Henriques_Torr_Vedaldi_2018}, and FC100 \cite{Oreshkin2018}.
\textit{mini}ImageNet and \textit{tiered}ImageNet are the subcollections of ImageNet \cite{Krizhevsky_Sutskever_Hinton_2017}, while CIFAR-FS and FC100 are derived from CIFAR100 \cite{Krizhevsky_2009}.
We adhere to the common practice data splits, consistent with \cite{chen2023semantic}, \cite{hao2023class}, where the data is mutually disjointly divided into the meta-training, meta-validation, and meta-test sets.

\subsection{Implementation Details}
\label{sec:implementation}
Our experiments are conducted under the $5$-way $1$-shot and $5$-shot settings.
We employ the Visformer-T \cite{Chen_Xie_Niu_Liu_Wei_Tian_2021} and Vit-S \cite{Dosovitskiy_Beyer_Kolesnikov_Weissenborn_Zhai_Unterthiner_Dehghani_Minderer_Heigold_Gelly_et_al} as backbone feature extractors, following the same pre-training and data augmentation strategies as in prior works \cite{hao2023class}, \cite{chen2023semantic}.
To derive comprehensive textual knowledge from class names, we employ the text encoder of CLIP \cite{Radford_Kim_Hallacy_Ramesh_Goh_Agarwal_Sastry_Askell_Mishkin_Clark_et_al}, which has undergone pre-training on extensive corpora.
We default to resizing the input image to 224$\times$224.
The number of learnable tokens in meta-class-specific prompts is 8.
The weight factor $\mu$ used in bidirectional knowledge permeation is set to 0.2.
The weight parameter $\gamma$ in the loss $\mathcal{L}_{total}$ is empirically set to 0.5.
Alternative selections for these parameters are validated in Sec. \ref{hyperparameters}.

BiKop framework is instantiated in PyTorch \cite{Paszke_Gross_Massa_Lerer_Bradbury_Chanan_Killeen_Lin_Gimelshein_Antiga_et} and executes for 100 epochs utilizing an NVIDIA RTX3090 GPU equipped with 24GB of memory.
During meta-training, the network is optimized using the AdamW optimizer\cite{Loshchilov} with a weight decay of 1e$-$5 and an initial learning rate of 2e$-$6.
Additionally, the learning rate for the BKP module is increased by a factor of 10, and for the SAD module, it is increased by a factor of 50.
During inference, a random sampling of 2,000 test episodes with 15 query images per class from the novel set is conducted, and the average accuracy along with a 95\% confidence interval is reported.

\subsection{Comparison with State-of-the-arts (SOTAs)}
\label{sec:Comparison}

We comprehensively compare BiKop with previous algorithms under $1$-shot and $5$-shot settings, where methods KTN \cite{Peng_Li_Zhang_Li_Qi_Tang_2019}, AM3 \cite{Chen_Rostamzadeh_Oreshkin_Pinheiro_2019}, and TRAML \cite{Li_Huang_Lan_Feng_Li_Wang_2020}, DeepBERT \cite{Yan_Bouraoui_Wang_Jameel_Schockaert_2021}, and SP \cite{chen2023semantic} similarly leverage textual information from class names.
% \vspace{\baselineskip}
% \noindent
% \textbf{\textit{mini}ImageNet.}
\subsubsection{\textit{mini}ImageNet.}
\label{compare_mini}
As shown in Table \ref{tab_1}, BiKop demonstrates a significant performance improvement compared to existing methods.
With the Visformer-T backbone, BiKop achieves a classification accuracy of 5.76\% higher than SP \cite{chen2023semantic}, which also incorporates textual information as prior knowledge.
When employing the meticulously pre-trained Vit-S backbone, we attain a performance improvement of 7.58\% compared to CPEA \cite{hao2023class}.
Under the $5$-shot setting, our approach achieves an obvious lead, \textit{i.e.,} 1.01\% with the ViT-S backbone.
Moreover, it is still on par with previous SOTAs when using the Visformer-T backbone.
%with an improvement up to 4.53%

\begin{table*}[t]
    \small
    \centering
    \caption{Comparison with prior approaches for the $5$-way $1$-shot and $5$-way $5$-shot settings on \textit{mini}ImageNet \cite{Vinyals_Blundell_Lillicrap_Kavukcuoglu_Wierstra_2016} and \textit{tiered}ImageNet \cite{Ren_Triantafillou_Ravi_Snell_Swersky_Tenenbaum_Larochelle_Zemel}. Methods in the top part do not involve textual information, the middle part introduces textual knowledge from class names, and the bottom part illustrates our methodology.}
    \resizebox{\linewidth}{!}{
        \begin{tabular}{lclcccc}
        \toprule
            & & & \multicolumn{2}{c}{\emph{mini}ImageNet 5-way} &  \multicolumn{2}{c}{\emph{tiered}ImageNet 5-way} \\
            Method  & Backbone & Params & 1-shot & 5-shot & 1-shot & 5-shot \\
        \midrule
            LEO \cite{Rusu_2018} & WRN-28-10 & 36.5M & 61.76$\pm$0.08 & 77.59$\pm$0.12 & 66.33$\pm$0.05 & 81.44$\pm$0.09 \\
            CC+rot \cite{Gidaris_2019} & WRN-28-10 & 36.5M & 62.93$\pm$0.45 & 79.87$\pm$0.33 & 70.53$\pm$0.51 & 84.98$\pm$0.36 \\
            Align \cite{Afrasiyabi_2019} & WRN-28-10 & 36.5M & 65.92$\pm$0.60 & 82.85$\pm$0.55 & 74.40$\pm$0.68 & 86.61$\pm$0.59 \\
            ProtoNet \cite{snell2017prototypical} & ResNet-12 & 12.5M & 62.29$\pm$0.33 & 79.46$\pm$0.48 & 68.25$\pm$0.23 & 84.01$\pm$0.56 \\
            MetaOptNet \cite{Lee_Maji_2019} & ResNet-12 & 12.5M & 62.64$\pm$0.61 & 78.63$\pm$0.46 & 65.99$\pm$0.72 & 81.56$\pm$0.53 \\
            Meta-Baseline \cite{Chen_Liu_Xu_2021} & ResNet-12 & 12.5M & 63.17$\pm$0.23 & 79.26$\pm$0.17 & 68.62$\pm$0.27 & 83.74$\pm$0.18\\
            DeepEMD \cite{Zhang_Cai_Lin_Shen_2020} & ResNet-12 & 12.5M & 65.91$\pm$0.82 & 82.41$\pm$0.56 & 71.16$\pm$0.87 & 86.03$\pm$0.58 \\
            Feat \cite{Ye_Hu_Zhan_Sha_2020} & ResNet-12 & 12.5M & 66.78$\pm$0.20 & 82.05$\pm$0.14 & 70.80$\pm$0.23 & 84.79$\pm$0.16 \\
            % IEPT \cite{Zhang_Huang_2021} & ResNet-12 & 12.5M & 67.05$\pm$0.44 & 82.90$\pm$0.30 & 72.24$\pm$0.50 & 86.73$\pm$0.34 \\
            TPMN \cite{Wu_Zhang_Zhang_Wu_2021} & ResNet-12 & 12.5M & 67.64$\pm$0.63 & 83.44$\pm$0.43 & 72.24$\pm$0.70 & 86.55$\pm$0.63\\
            SetFeat \cite{Afrasiyabi_2022} & ResNet-12 & 12.5M & 68.32$\pm$0.62 & 82.71$\pm$0.46 & 73.63$\pm$0.88 & 87.59$\pm$0.57 \\
            FGFL \cite{Cheng_Yang_Zhou_Guo_Wen_Query} & ResNet-12 & 12.5M & 69.14$\pm$0.80 & 86.01$\pm$0.62 & 73.21$\pm$0.88 & 87.21$\pm$0.61 \\
            SUN \cite{Dong_Zhou_Yan_Zuo} & Visformer-S & 12.4M & 67.80$\pm$0.45 & 83.25$\pm$0.30 & 72.99$\pm$0.50 & 86.74$\pm$0.33 \\
            FewTURE \cite{hiller2022rethinking} & ViT-S & 22.0M & 68.02$\pm$0.88 & 84.51$\pm$0.53 & 72.96$\pm$0.92 & 86.43$\pm$0.67 \\
            CPEA \cite{hao2023class} & ViT-S & 22.0M & \textbf{71.97$\pm$0.65} & \textbf{87.06$\pm$0.38} & \textbf{76.93$\pm$0.70} & \textbf{90.12$\pm$0.45} \\
            \midrule
            KTN \cite{Peng_Li_Zhang_Li_Qi_Tang_2019} & ResNet-12 & 12.5M & 61.42$\pm$0.72 & 74.16$\pm$0.56 & - & - \\
            AM3 \cite{Chen_Rostamzadeh_2019} & ResNet-12 & 12.5M & 65.30$\pm$0.49 & 78.10$\pm$0.36 & 69.08$\pm$0.47 & 82.58$\pm$0.31 \\
            TRAML \cite{Li_Huang_Lan_Feng_Li_Wang_2020} & ResNet-12 & 12.5M & 67.10$\pm$0.52 & 79.54$\pm$0.60 & - & - \\
            DeepEMD-BERT \cite{Yan_Bouraoui_2021} & ResNet-12 & 12.5M & 67.03$\pm$0.79 & \textbf{83.68$\pm$0.65} & 73.76$\pm$0.72 & 87.51$\pm$0.75\\
            SP \cite{chen2023semantic} & Visformer-T & 10.0M & \textbf{72.31$\pm$0.40} & 83.42$\pm$0.30 & \textbf{78.03$\pm$0.46} & \textbf{88.55$\pm$0.32}\\
            \midrule
            Pre-train (Ours) & Visformer-T & 10.0M & 67.36$\pm$0.46 & 81.25$\pm$0.40 & 72.55$\pm$0.47 & 86.59$\pm$0.29 \\
            \textbf{BiKop} (Ours)  & Visformer-T & 10.0M & 78.07$\pm$0.36 & 83.53$\pm$0.29 &  80.29$\pm$0.43 & 88.26$\pm$0.32 \\
            \textbf{BiKop} (Ours)  & ViT-S & 22.0M & \textbf{79.55$\pm$0.50} & \textbf{88.07$\pm$0.37} & \textbf{82.98$\pm$0.63} & \textbf{90.33$\pm$0.44} \\
        \bottomrule
        \end{tabular}}
    \label{tab_1}
    \vspace{-0.4cm}
\end{table*}

% \vspace{\baselineskip}
% \noindent
% \textbf{\textit{tiered}ImageNet.}
\subsubsection{\textit{tiered}ImageNet.}
\label{compare_tiered}
By adopting the same backbone ViT-S, BiKop outperforms the previous best method CPEA \cite{hao2023class} with advantages of 6.05\% and 0.21\% in $1$-shot and $5$-shot settings, respectively. 
When the backbone of Visformer-T is utilized, BiKop obtains an improvement of 2.26\% under the $1$-shot setting.

\begin{table*}[t]
    \small
    \centering
    \caption{Comparison with peers for the $5$-way $1$-shot and $5$-way $5$-shot settings on CIFAR-FS \cite{Bertinetto_Henriques_Torr_Vedaldi_2018} and FC100 \cite{Oreshkin2018}.}
    \resizebox{\linewidth}{!}{
        \begin{tabular}{lclcccc}
        \toprule
            & & & \multicolumn{2}{c}{CIFAR-FS 5-way} &  \multicolumn{2}{c}{FC100 5-way} \\
            Method  & Backbone & Params & 1-shot & 5-shot & 1-shot & 5-shot \\
        \midrule
            PN+rot \cite{Gidaris_2019} & WRN-28-10 & 36.5M & 69.55$\pm$0.34 & 82.34$\pm$0.24 & - & - \\
            Align \cite{Afrasiyabi_2019} & WRN-28-10 & 36.5M & - & - & 45.83$\pm$0.48 & 59.74$\pm$0.56 \\
            % PSST \cite{Chen_Ge_Zhan_Huang_Wang_2021} & WRN-28-10 & 36.5M &77.02$\pm$0.38 & 88.45$\pm$0.35 & - & - \\
            Meta-QDA \cite{Zhang_Meng_Gouk_Hospedales_2021} & WRN-28-10 & 36.5M &75.83$\pm$0.88 & 88.79$\pm$0.75 & - & - \\
            ProtoNet \cite{snell2017prototypical} & ResNet-12 & 12.5M & 72.20$\pm$0.70 & 83.50$\pm$0.50 & 37.50$\pm$0.60 & 52.50$\pm$0.60 \\
            MetaOptNet \cite{Lee_Maji_2019} & ResNet-12 & 12.5M & 72.60$\pm$0.70 & 84.30$\pm$0.50 & 41.10$\pm$0.60 & 55.50$\pm$0.60 \\
            % MABAS \cite{Kim_Kim_Kim_2020} & ResNet-12 & 12.5M & 73.51$\pm$0.92 & 85.49$\pm$0.68 & 42.31$\pm$0.75 & 57.56$\pm$0.78 \\
            Distill \cite{Tian_Wang_Krishnan_Tenenbaum_Isola_2020} & ResNet-12 & 12.5M & 73.90$\pm$0.80 & 86.90$\pm$0.50 & 44.60$\pm$0.70 & 60.90$\pm$0.60 \\ 
            BML \cite{Zhou_Qiu_Xie_Wu_Zhang_2021} & ResNet-12 & 12.5M & 73.45$\pm$0.47 & 88.04$\pm$0.33 & - & - \\ 
            CG \cite{Gao_Wu_Jia_Harandi_2021} & ResNet-12 & 12.5M & 73.00$\pm$0.70 & 85.80$\pm$0.50 & - & - \\ 
            % Meta-NVG \cite{Zhang_Ding_Lin_Li_Wang_Shen_2021} & ResNet-12 & 12.5M & 74.63$\pm$0.91 & 86.45$\pm$0.59 & 46.40$\pm$0.81 & 61.33$\pm$0.71 \\ 
            TPMN \cite{Wu_Zhang_Zhang_Wu_2021} & ResNet-12 & 12.5M & 75.50$\pm$0.90 & 87.20$\pm$0.60 & 46.93$\pm$0.71 & 63.26$\pm$0.74\\
            MixFSL \cite{Afrasiyabi_Lalonde_Gagne_2021} & ResNet-12 & 12.5M & - & - & 44.89$\pm$0.63 & 60.70$\pm$0.60\\
            infoPatch \cite{Liu_Fu_Xu_Yang_Li_Wang_Zhang_2022} & ResNet-12 & 12.5M & - & - & 43.80$\pm$0.40 & 58.00$\pm$0.40 \\
            SUN \cite{Dong_Zhou_Yan_Zuo} & Visformer-S & 12.4M & 78.37$\pm$0.46 & 88.84$\pm$0.32 & - & - \\
            SP \cite{chen2023semantic} & Visformer-T & 10.0M & \textbf{82.18$\pm$0.40} & 88.24$\pm$0.32 & \textbf{48.53$\pm$0.38} & 61.55$\pm$0.41\\
            FewTURE \cite{hiller2022rethinking} & ViT-S & 22.0M & 76.10$\pm$0.88 & 86.14$\pm$0.64 & 46.20$\pm$0.79 & 63.14$\pm$0.73 \\
            CPEA \cite{hao2023class} & ViT-S & 22.0M & 77.82$\pm$0.66 & \textbf{88.98$\pm$0.45} & 47.24$\pm$0.58 & \textbf{65.02$\pm$0.60} \\
            \midrule
            Pre-train (Ours) & Visformer-T & 10.0M & 72.24$\pm$0.49 & 85.76$\pm$0.36 & 43.56$\pm$0.39 & 59.47$\pm$0.39 \\
            \textbf{BiKop} (Ours)  & Visformer-T & 10.0M & \textbf{84.83$\pm$0.35} & 88.89$\pm$0.31 & 52.14$\pm$0.39 & 62.41$\pm$0.40 \\
            \textbf{BiKop} (Ours)  & ViT-S & 10.0M & 84.19$\pm$0.58 & \textbf{89.70$\pm$0.47} & \textbf{53.25$\pm$0.53} & \textbf{65.07$\pm$0.56} \\
        \bottomrule
        \end{tabular}}
    \label{tab_2}
    \vspace{-0.3cm}
\end{table*}

% \vspace{\baselineskip}
% \noindent
% \textbf{CIFAR-FS.}
\subsubsection{CIFAR-FS.}
\label{compare_CIFAR}
Compared to the competitive SP \cite{chen2023semantic}, BiKop achieves 2.65\% and 0.65\% higher accuracies in the $1$-shot and $5$-shot settings, respectively.
With the backbone of ViT-S, the proposed BiKop outperforms the preeminent CPEA \cite{hao2023class} with a performance improvement of 6.37\% in $1$-shot and 0.72\% in $5$-shot.

% \vspace{\baselineskip}
% \noindent
% \textbf{FC100.}
\subsubsection{FC100.}
\label{compare_FC100}
For the more challenging dataset FC100, BiKop again sets a new SOTA performance across various settings.
Based on the backbone of ViT-S, BiKop obtains a significant improvement of 6.01\% in the $1$-shot scenario and gets a similar result to CPEA \cite{hao2023class} in the $5$-shot setting.
Meanwhile, with the backbone of Visformer-T, there is an increase of 3.61\% ($1$-shot) and 0.86\% ($5$-shot).

\vspace{\baselineskip}
\noindent
In summary, the model's performance has seen a tremendous improvement compared to the results of pre-training.
More importantly, BiKop demonstrates a substantial performance lead in the scenario of extremely limited samples, \textit{i.e.,} $1$-shot setting, affirming that our strategy takes mutual advantage of the textual and visual knowledge, establishing a more robust sparse feature representation.
In the $5$-shot scenario, we obtain a slight performance improvement as the richer visual features of support samples dominate over the complementarity of textual knowledge.
For the subsequent experiments, we adopt Visformer-T as the default backbone.

\begin{table}[t]
  \centering
  \caption{Ablation study under the $1$-shot setting. i$\rightarrow$t and i$\leftarrow$t represent the unidirectional permeation from visual to textual modalities only and the opposite direction.}
  \setlength{\tabcolsep}{1.55mm}{
    \begin{tabular}{cccc|cccc}
    \hline
    \multirow{2}[2]{*}{Meta-Prompt} & \multicolumn{2}{c}{BKP} & \multirow{2}[2]{*}{SAD} & \multirow{2}[2]{*}{\textit{mini}} & \multirow{2}[2]{*}{\textit{tired}} & \multirow{2}[2]{*}{CIFAR-FS} & \multirow{2}[2]{*}{FC100} \\
          & i$\rightarrow$t   & i$\leftarrow$t   &       &       &       &       &  \\
    \hline
    \ding{55} & \ding{55} & \ding{55} & \ding{55} & 71.88 & 76.51 & 81.34 & 47.71  \\
    \ding{52} & \ding{55} & \ding{55} & \ding{55} & 76.36 & 78.06 & 82.50 & 50.08 \\
    \ding{52} & \ding{55} & \ding{52} & \ding{55} & 77.07 & 79.15 & 84.03 & 50.48 \\
    \ding{52} & \ding{52} & \ding{55} & \ding{55} & 76.82 & 78.31 & 82.97 & 50.96 \\
    \ding{52} & \ding{52} & \ding{52} & \ding{55} & 77.23 & 79.57 & 84.43 & 51.21 \\
    \ding{55} & \ding{52} & \ding{52} & \ding{52} & 76.65 & 78.21 & 84.14 & 50.69 \\
    \ding{52} & \ding{52} & \ding{52} & \ding{52} & \textbf{78.07} & \textbf{80.29} & \textbf{84.83} & \textbf{52.14} \\
    \hline
    \end{tabular}}%
  \label{tab_3}%
\end{table}%

\subsection{Ablation Study}
\label{sec:Ablation}

\begin{figure}[t]
  \begin{minipage}[t]{0.48\textwidth}
    \centering
    \includegraphics[height=4.2cm]{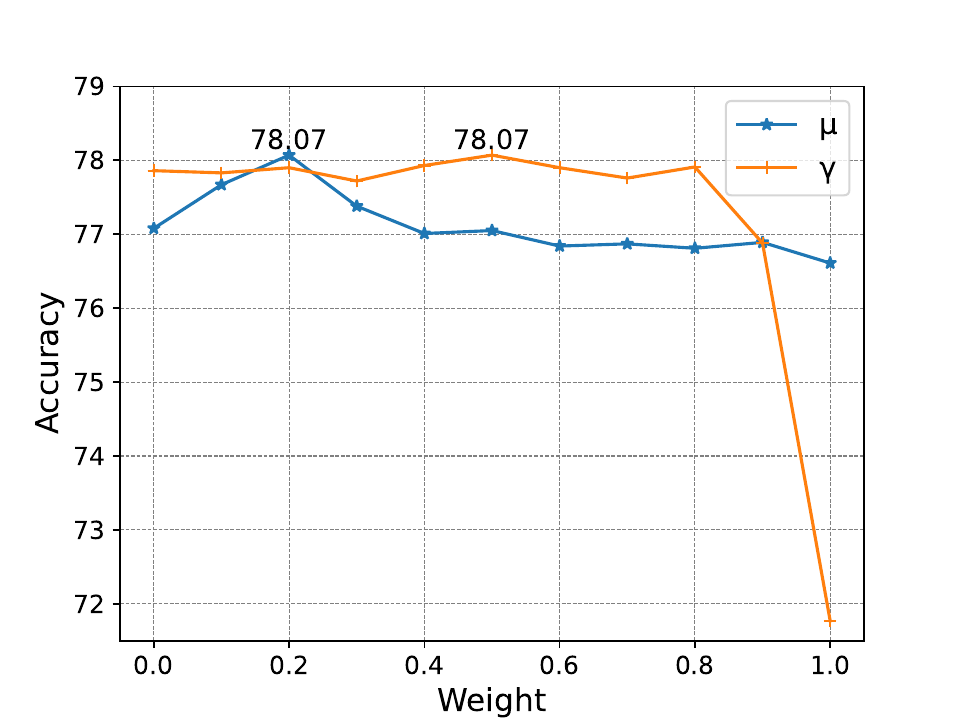}
    \captionsetup{width=0.9\linewidth, justification=justified}
    \caption{Effect of weight coefficients $\mu$ in the BKP moddule and $\gamma$ in the overall loss on \textit{mimi}ImageNet under $1$-shot setting.}
    \label{fig_4}
  \end{minipage}%
  \begin{minipage}[t]{0.52\textwidth}
    \centering
    \includegraphics[height=4.2cm]{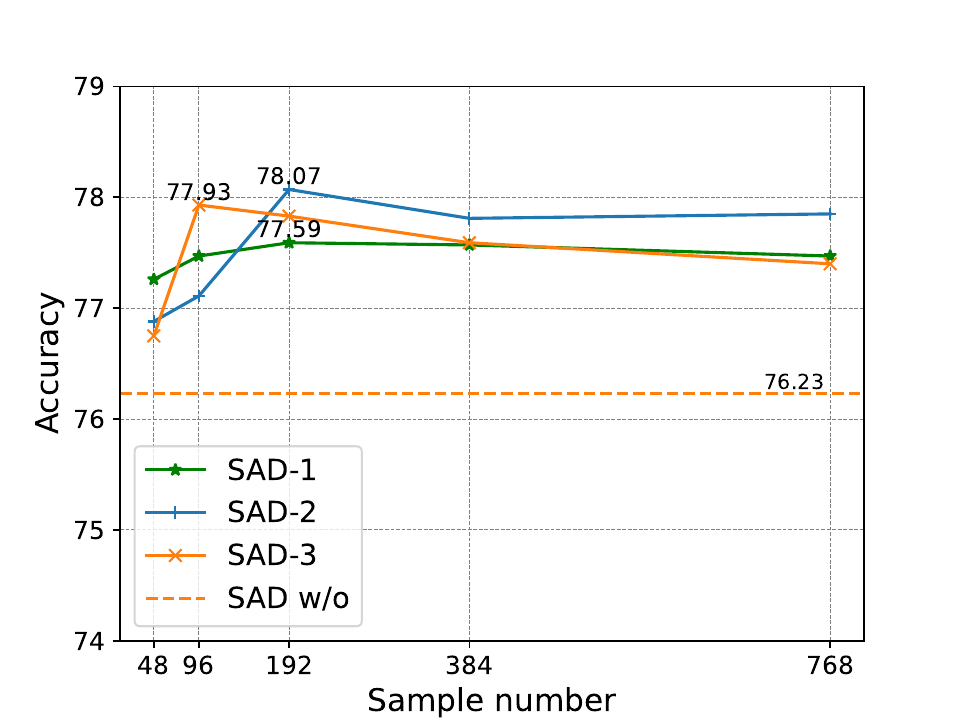}
    \captionsetup{width=0.9\linewidth, justification=justified}
    \caption{Effect of sampling times $m$ and layer number of MLP block $\mathcal{D}(\cdot)$ in the SAD module on \textit{mimi}ImageNet under $1$-shot setting.}
    \label{fig_5}
  \end{minipage}
\end{figure}

\subsubsection{Efficacy of different components in BiKop.}
\label{components}
The ablation study results on key components of BiKop are shown in Table. \ref{tab_3}.
First, we use the original class name embedding as textual knowledge and remove the proposed BKP and SAD modules, considering this as our baseline.
Then, we get significant accuracy improvement (2.39\% on average over four datasets) when incrementally introducing the meta-class-specific prompt.
To validate the effectiveness of BKP, we transform it into two unidirectional knowledge permeations, \textit{i.e.,} from text to vision and vice versa.
Observing the results, bidirectional knowledge permeation significantly outperforms unidirectional permeation, suggesting that enhancing the generalization of visual knowledge and the individual diversity of textual knowledge is advantageous to establishing a robust joint feature representation.
Finally, by incorporating the SAD module, the performance is further enhanced.

\begin{table}[t]
  \centering
  \caption{Impact of different implementations of BKP under the $1$-shot setting.}
  \setlength{\tabcolsep}{1.55mm}{
    \begin{tabular}{c|cccc}
    \hline
          & Mini  & Tired & CIFAR-FS & FC100 \\
    \hline
    Dot product & 59.91 & 67.63 & 64.38 & 38.02 \\
    Addition & 75.91 & 79.44 & 81.83 & 47.58 \\
    Concatenation & 74.36 & 78.24 & 82.76 & 49.02 \\
    Cross-attention (ours) & \textbf{78.07} & \textbf{80.29} & \textbf{84.83} & \textbf{52.14} \\
    \hline
    \end{tabular}}
  \label{tab_4}%
\end{table}%

\begin{figure}[t]
  \begin{minipage}[t]{0.48\textwidth}
    \centering
    \includegraphics[height=4.2cm]{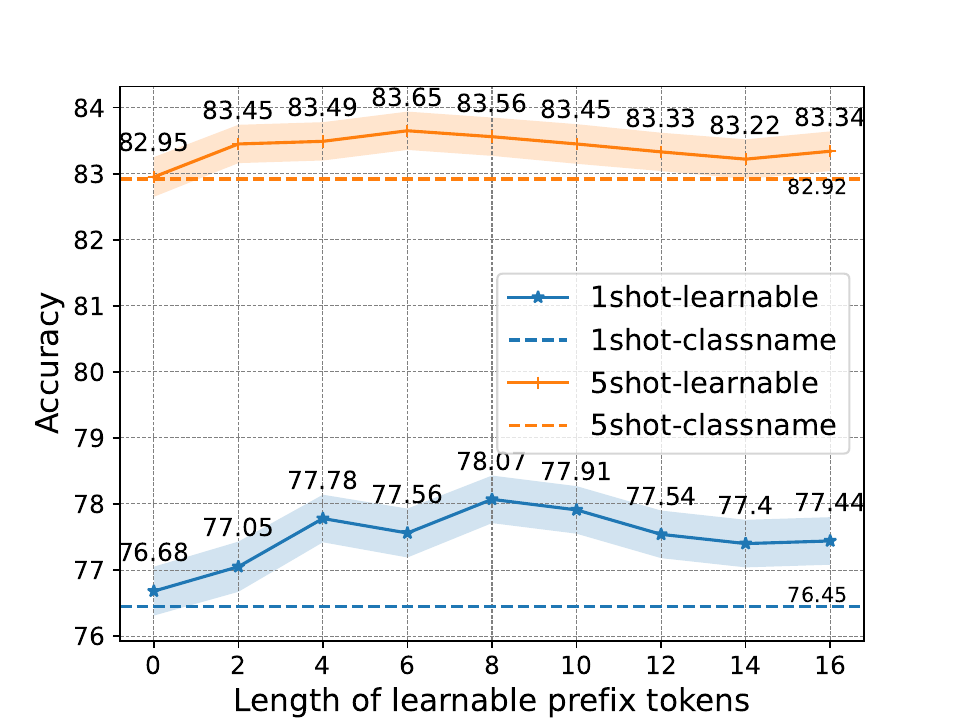}
    \captionsetup{width=0.9\linewidth, justification=justified}
    \caption{Designs for the meta-class-specific prompt on \textit{mini}ImageNet.}
    \label{fig_6}
  \end{minipage}%
  \begin{minipage}[t]{0.52\textwidth}
    \centering
    \includegraphics[height=4.2cm]{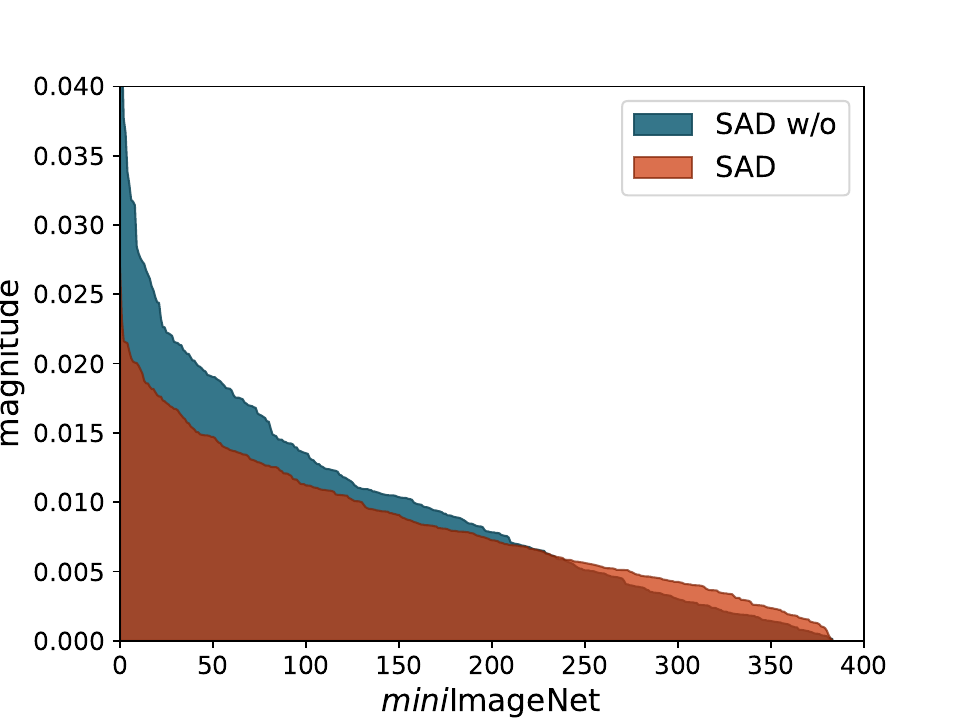}
    \captionsetup{width=0.9\linewidth, justification=justified}
    \caption{Visualization results on Mean Magnitude features of Channels (MMC).}
    \label{fig_7}
  \end{minipage}
\end{figure}

\subsubsection{Influence of hyperparameters.}
\label{hyperparameters}
From Fig. \ref{fig_4}, it can be observed that as $\mu$ increases, the performance improves initially but then experiences a significant decrease.
We posit that an excessively large $\mu$ leads to the suppression of intrinsic modality knowledge, which might be detrimental to establishing a robust joint feature representation.
Meanwhile, a sharp decline occurred in performance when $\gamma$ becomes too large since an over-high proportion of adversarial loss might bring catastrophic optimization of the model.
Additionally, for the choice of sampling times $m$ and the layer number of MLP block $\mathcal{D}(\cdot)$ in the SAD module, we compare the corresponding performance in Fig. \ref{fig_5}.
We employed a two-layer MLP block and set the sampling times $m$ to 192, thus achieving optimal performance.

\begin{figure}[t]
    \centering
    \includegraphics[width=0.85\textwidth]{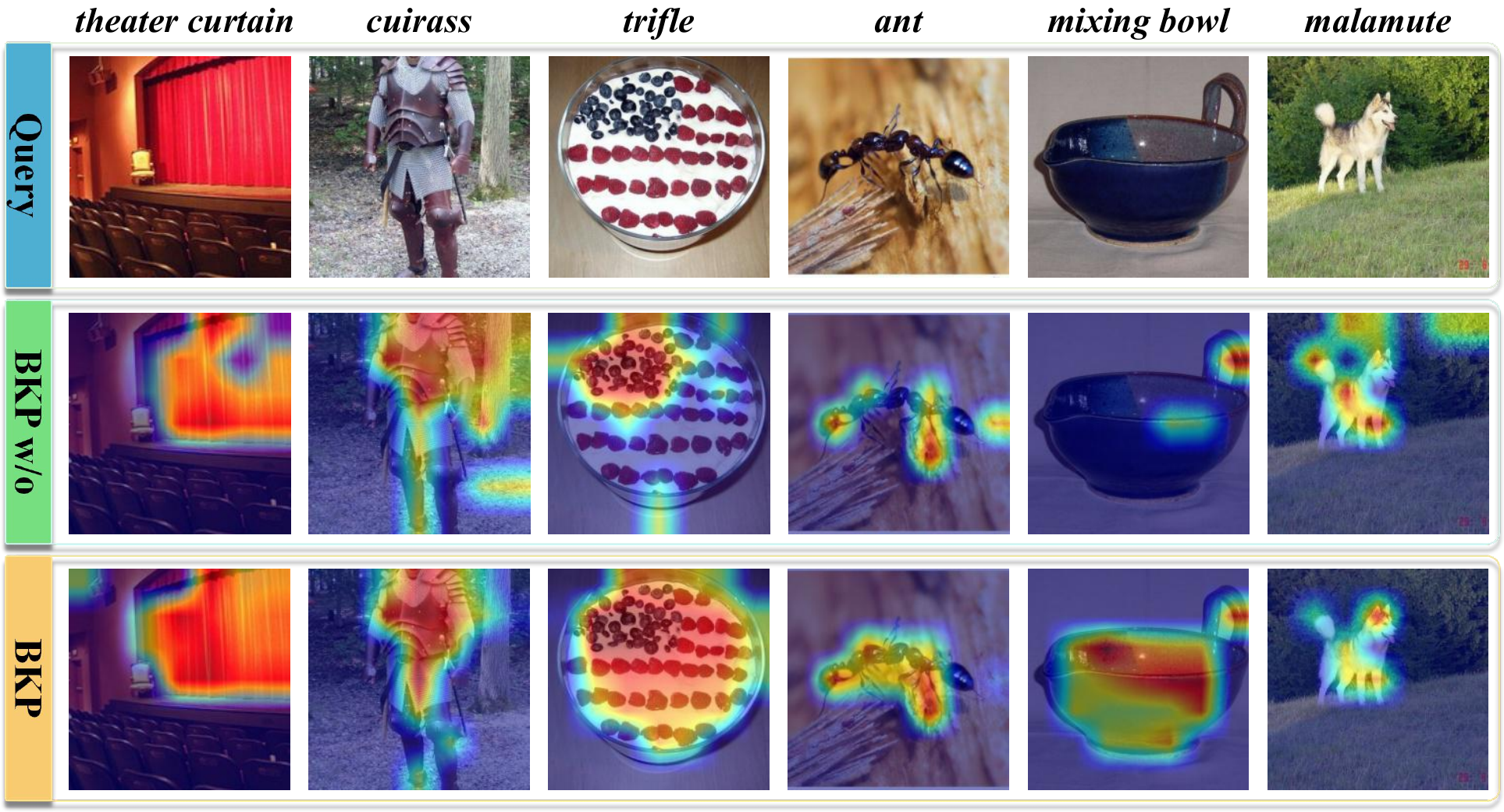}
    \caption{Qualitative comparison of with and without BKP.} 
    \label{fig_8}
\end{figure}

\subsubsection{Implementation of BKP.}
\label{Implementation of BKP}
We investigate the impact of different implementation strategies for BKP.
Let $GAP(\cdot)$ stands for the Global Average Pooling function, $\bm{Z}_{l}^{s} \in \mathbb{R}^{M \times C_{d}}$ and $\mathbf{T}\in \mathbb{R}^{1 \times C_{d}}$ represent the visual feature and textual knowledge, respectively.
``Dot product" can be formulated as:
\begin{equation}
\label{eqt_16}
\bm{\hat{Z}}_{l}^{s}=\bm{Z}_{l}^{s}\cdot \mathbf{T}, \ \ \ \mathbf{\hat{T}}=\mathbf{T}\cdot GAP(\bm{Z}_{l}^{s}),
\end{equation}
``Addition'' can be denoted by:
\begin{equation}
\label{eqt_17}
\bm{\hat{Z}}_{l}^{s}=\bm{Z}_{l}^{s} + \mathbf{T}, \ \ \ \mathbf{\hat{T}}=\mathbf{T} + GAP(\bm{Z}_{l}^{s}),
\end{equation}
``Concatenation'' concatenates $\bm{Z}_{l}^{s}$ and $\mathbf{T}$ along the spatial dimensions, similar to the ``spatial interaction'' in SP \cite{chen2023semantic}.
As shown in Table \ref{tab_4}, our design better harnesses the mutual advantages of both the text and knowledge modalities, thanks to the powerful global information interaction capability of the cross-attention mechanism.

\subsubsection{Design of the meta-class-specific prompt.}
\label{meta-class-specific prompt}
We explore the impact of different prompt design strategies, including the original class name embedding (dashed lines) and various sizes of the meta-class-specific prompt, \textit{i.e.,} the number of learnable tokens.
The $1$-shot and $5$-shot performance is reported in Fig. \ref{fig_6}, yielding two observations:
First, compared to the class name embeddings, introducing the meta-class-specific prompt with learnable tokens undergoes a sharp increase in accuracy. 
This validates that our designed prompts better adapt to the meta-learning paradigm.
Second, as the size of the prompt increases, the performance exhibits a gradual improvement, followed by a slight decline.
The underlying cause can be deduced to be two-fold.
On the one hand, increasing the size of the prompt enhances the discrimination of textual knowledge between the meta-classes.
On the other hand, an oversized prompt (numerous learnable parameters) might potentially result in overfitting of the model.
Therefore, we set the number of learnable tokens to 8 to achieve an overall optimal performance.

\subsection{Visualization and Analysis}
\label{sec:Visualization}
Luo \textit{et al.} \cite{Luo_Xu_Xu_2022} indicated that channel importance is closely related to model bias, impacting the generalization of FSL. 
Mean Magnitude of Channels (MMC) can reveal channels' responses, high-performing few-shot methods may exhibit a more uniform MMC curve on the test set.
To this end, we depict the MMC over the test set of \textit{mini}ImageNet.
In Fig. \ref{fig_7}, the channel magnitude becomes more uniform after applying the SAD module, which validates that the proposed SAD module effectively alleviates the suppression on potential novel class semantics, \textit{i.e.,} the issue of model bias.

Furthermore, we visualize the spatial attention map of the output feature in Fig. \ref{fig_8}, where the red regions indicate higher attention values.
We notice that the BKP module assists the model in adapting attention to the objects responsible for classification, utilizing a classifier built with only a few support images. 
This underscores its efficacy in establishing the robust joint feature representation.

\section{Conclusion and Discussion}
\label{sec:Conclusion}
In this paper, we proposed a novel BiKop framework for FSL.
BiKop extends class names to meta-class-specific prompts and incorporates a bidirectional knowledge permeation (BKP) module, fully leveraging the mutual advantages of textual and visual knowledge. 
This effectively mitigates the problem of sparse feature representation collapse.
In addition, considering the model bias exacerbated by base-class-relevant information, we devised the Semantic Adversarial Disentanglement (SAD) module to loosen the suppression of potential novel-class semantics during training, further enhancing the model's generalization.
Experimental results on four datasets show that BKP performs much better than previous methods.
More in-depth ablation studies validate that our designs respond to the advocated motivation.

\bibliographystyle{splncs04}
\bibliography{ref}
\end{document}